\documentclass[letterpaper, 10 pt, conference]{ieeeconf}  

\IEEEoverridecommandlockouts                              
\usepackage{graphicx}                                                         
\usepackage{amsmath} 
\usepackage[caption=false, font=footnotesize]{subfig}
\usepackage{amssymb}
\usepackage{multirow}
\usepackage{multicol}
\usepackage{verbatim}
\usepackage{dsfont}
\usepackage{makecell}

\usepackage{array}
\newcolumntype{C}[1]{>{\centering\arraybackslash}p{#1}}

\newcommand{\hlineplus}{\Xhline{2\arrayrulewidth}}

\title{\LARGE \bf
Semantic Deep Intermodal Feature Transfer:\\ Transferring Feature Descriptors Between Imaging Modalities
}

\author{Sebastian P. Kleinschmidt$^{1}$ and Bernardo Wagner$^{1}$
\thanks{$^{1}$The authors are with the Institute of Systems Engineering - Real Time Systems Group,
        Leibniz Universit{\"a}t Hannover, 30167, Germany
        {\tt\small \{kleinschmidt,wagner\}@rts.uni-hannover.de}}%
}

\begin{document}

\maketitle
\thispagestyle{empty}
\pagestyle{empty}

\begin{abstract}
Under difficult environmental conditions, the view of RGB cameras may be restricted by fog, dust or difficult lighting situations. 
Because thermal cameras visualize thermal radiation, they are not subject to the same limitations as RGB cameras. However, because RGB and thermal imaging differ significantly in appearance, common, state-of-the-art feature descriptors are unsuitable for intermodal feature matching between these imaging modalities. As a consequence, visual maps created with an RGB camera can currently not be used for localization using a thermal camera.
In this paper, we introduce the \textit{Semantic Deep Intermodal Feature Transfer} (Se-DIFT), an approach for transferring image feature descriptors from the visual to the thermal spectrum and vice versa. For this purpose, we predict potential feature appearance in varying imaging modalities using a deep convolutional encoder-decoder architecture in combination with a global feature vector. Since the representation of a thermal image is not only affected by features which can be extracted from an RGB image, we introduce the global feature vector which augments the auto encoder's coding. The global feature vector contains additional information about the thermal history of a scene which is automatically extracted from external data sources. By augmenting the encoder's coding, we decrease the L1 error of the prediction by more than $7 \%$ compared to the prediction of a traditional U-Net architecture. To evaluate our approach, we match image feature descriptors detected in RGB and thermal images using Se-DIFT. Subsequently, we make a competitive comparison on the intermodal transferability of \textit{SIFT}, \textit{SURF}, and \textit{ORB} features using our approach. As shown in the evaluation, feature matching using Se-DIFT results in a drastically increased area under curve (AUC) and decreased equal error rates (EER) of the receiver operator curves (ROC) for intermodal feature matching compared to a direct intermodal matching as well as matching based on intermodal predictions of other network architectures.
\end{abstract}

\section{INTRODUCTION}
\begin{figure}[t]
      \centering
      \includegraphics[scale=0.5]{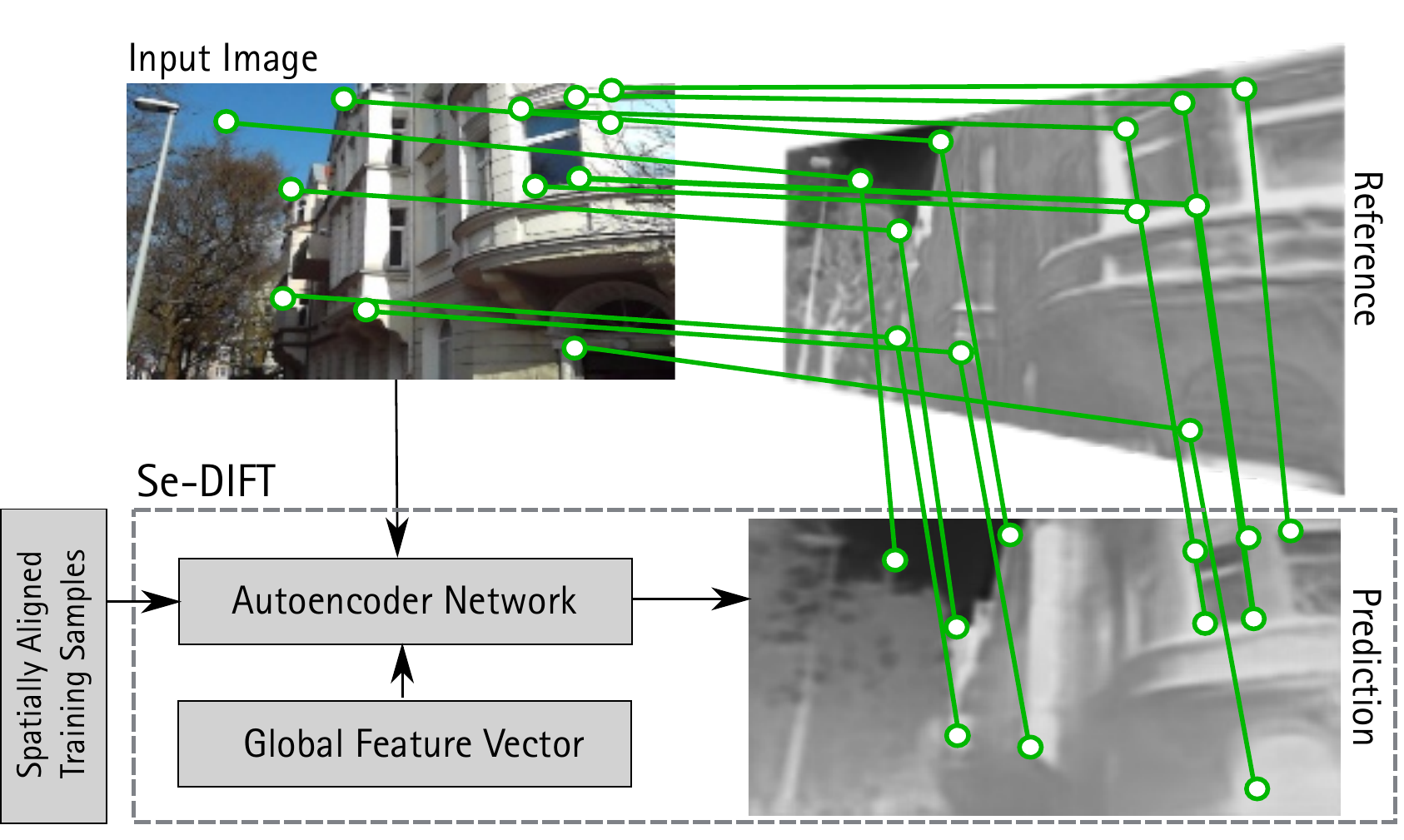}
      \caption{The Se-DIFT approach can be applied to transfer image feature descriptors between the visual and thermal using a convolutional encoder-decoder neural network architecture predicting the appearance of a feature in the target imaging modality.}
      \label{sediftconcept}
   \end{figure}

\noindent Image feature matching is an essential component of many computer vision applications such as image alignment, object recognition, tracking, 3D reconstruction, and robot navigation.
Modern feature detectors and descriptors such as \textit{ORB} \cite{orbfeatures}, \textit{SIFT} \cite{SIFT}, and \textit{SURF} \cite{SURF} have proven to identify feature correspondences robustly despite a variety of image transformations like translations and rotations. In presence of environmental variations as changing lighting or reduced visibility due to smoke or dust, their performance is limited. Some of these limitations can be compensated using varying imaging modalities such as thermal cameras, which are independent of changing lighting conditions, only visualizing thermal radiation. Especially for search and rescue robotics, thermal imaging has proven to be particularly useful in terms of victim localization and by providing clear images even in the presence of smoke.
Due to the fact that visual and thermal imaging drastically distinguish in appearance, visualizing different regions of the electromagnetic spectrum, modern state-of-the-art feature descriptors have shown to be inappropriate for intermodal feature matching. In case of thermal imaging, a low spatial resolution, significant Gaussian noise as well as more uniformly image regions further increase the difficulty of correspondence identification~\cite{Lin2001}.
The \textit{Semantic Deep Intermodal Feature Transfer} (Se-DIFT) presented in this paper enables the identification of image correspondences across imaging modalities using traditional feature descriptors. Therefore, Se-DIFT transforms direct intermodal feature matching into an indirect matching by predicting potential feature appearances intermodally before feature descriptors are computed. Using this approach, current state-of-the-art feature matching approaches become applicable for intermodal correspondence identification. The basic structure of Se-DIFT is visualized in Fig \ref{sediftconcept}.

From the author's perspective, Se-DIFT is the first approach which successfully transfers state-of-the-art image feature descriptors between the visual and thermal spectrum. Therefore, the main contributions of this paper are:
\begin{itemize}
\item Intermodal feature descriptor transfer between the thermal and visual spectrum by predicting the appearance of a feature in the target imaging modality.
\item An augmented convolutional encoder-decoder neural network architecture for predicting intermodal feature appearances using local image features as well as a global feature vector which augments the encoder's coding.
\end{itemize}
Even if Se-DIFT is exclusively applied to the problem of RGB and thermal feature matching in the scope of this paper, the idea of Se-DIFT is probably applicable for intermodal feature matching of other imaging modalities as well. Accordingly, Se-DIFT has the potential to be used for applications such as intermodal image-based localization as well as multimodal structure from motion and further expands the interchangeability of visual maps across robots which are equipped with different imaging sensors. 

\section{RELATED WORK}
\noindent In this section, we will give a brief overview of relevant publications in the area of multimodal imaging. Furthermore, we will introduce related publications in the field of deep learning-based image colorization and image-to-image translation.
\subsection{Multimodal Imaging}
\noindent Establishing spatial relations between images of different modalities is relevant for a variety of applications. Multimodal image fusion algorithms have shown notable achievements in improving the clinical accuracy of decisions based on medical images \cite{Pappachen2014}. Due to the large number of methods for medical image fusion which are based on knowledge, wavelets, fuzzy logic, artificial neural networks or morphology operators, the reader is referred to the surveys presented in \cite{Pappachen2014} and \cite{James2015}.

In contrast to imaging for robotic applications, medical imaging has to deal with a significantly smaller number of different objects, mainly from a known anatomic atlas. In general, the object of investigation is known apriori, and only a transformation for image alignment is needed for most applications. Due to the more sophisticated image acquisition procedure for many medical applications, images are usually captured in controlled environments with a limited number of unknowns. Accordingly, most medical image registration algorithms do not depend on sparse correspondence identification using image features. Other applications, in contrast, have to deal with a large number of different and potentially unknown objects and often have no initial information about their environment or about the camera pose. Image feature matching methods can be used to determine unknown camera poses which are needed for applications such as image-based localization, structure from motion or visual mapping. Nevertheless, there is only a limited number of publications regarding the use of multimodal image features in robotic applications.

The first analysis of the spatial and statistical distribution of sparse multimodal image features in a robotic context has been presented in \cite{Kleinschmidt2017}. Therefore, the authors created spatially aligning images of different imaging modalities according to \cite{Kleinschmidt2018A}, followed by detecting sparse image features using a Harris corner detector. The features have then been categorized into uni-, bi-, and multimodal image features using a probabilistic fusion approach. The authors conclude, that the number of intermodally transferable bi- and multimodal image features is relatively small compared to the number of unimodal image features which can be detected in the individual imaging modality. Another drawback is the fact that features, especially in the thermal spectrum, appeared unequally distributed over the image in indoor environments.

In \cite{Kleinschmidt2018B} the first approach to create visual multimodal maps using sparse image features has been introduced: The authors present an approach for visual odometry using sparse image features of different imaging modalities to create a multimodal map. Even if the pose estimation based on the RGB image has shown to be the most stable, image features of additional imaging modalities could be mapped and used to overcome partial sensor failures of the RGB camera as caused by smoke. Nevertheless, because of the missing intermodal transferability of image feature descriptors, the resulting mapped features could only be used to match features extracted from the same modality.

\subsection{Deep Learning-based Image-to-Image Translation}
\noindent The authors of \cite{px2px} define the task of \textit{translating one possible representation of a scene into another, given sufficient training data} as \textit{automatic image-to-image translation}. The first neural network based image colorization approach was presented in \cite{cheng2015}. The proposed approach uses a deep neural network for image colorization by formulating image colorization as a regression problem, predicting chrominance values. Patch- (\textit{low-level}), DAISY (\textit{mid-level}) and semantic features (\textit{high-level classification}) are extracted and used as input features for the network. In contrast to the deep neural network used in \cite{cheng2015}, in \cite{Iizuka2016} the authors present an end-to-end convolutional neural network architecture that jointly learns global and local features of an image for an user-intervention-free image colorization. Whereas global image features are used to determine semantic information (e.g., if an image has been taken indoor or outdoor or if it has been taken at day or night), local features are used to determine the object or local texture at a given location. Both kinds of features are fused to predict the chrominance of the uncolored input image. Like \cite{Iizuka2016}, \cite{Larsson2016} and \cite{Zhang2016} are using convolutional neural networks for image colorization: Instead of predicting a single color for an image pixel, the work presented in \cite{Larsson2016} predicts a color histogram. Their network architecture is based on the 16-layer configuration of the VGG16 network \cite{vgg16} which has been modified to operate on grayscale images. Additionally, they discard the final classification layer. A hypercolumn descriptor is extracted for each pixel by concatenating the features at its spatial location in all layers, followed by a fully connected layer used for histogram prediction. In \cite{Zhang2016}, class-rebalancing is used at training time to increase the diversity of colors in the result. The approach is evaluated using a colorized Turing test, in which human participants had to choose between generated and ground truth color images to proof the plausibility of their results. A more general network architecture for automatic image-to-image translation is presented in \cite{px2px}. Therefore they use a conditional Generative Adversarial Network (cGAN) which learns a cost function for different translation tasks such as day to night or cityscape labels to photos predictions. They also show example results for the translation of thermal to RGB images. In contrast to the approach presented in this paper, the authors of \cite{px2px} do not consider external information as input to their network neither they provide any additional information to quantify the performance of their Thermal-to-RGB prediction. Finally, whereas \cite{px2px} merely provide RGB predictions based on thermal images, this paper provides predictions for both imaging modalities.

\section{PROPOSED APPROACH}
\noindent The \textit{Semantic Deep Intermodal Feature Transfer} (Se-DIFT) enables the identification of image correspondences across imaging modalities using traditional feature descriptors. For this reason, we first predict the appearance of a scene in a different imaging modality to subsequently compute traditional image feature descriptors. Motivated by the recent success in deep learning approaches for image colorization and image-to-image translation, we designed a symmetric convolutional encoder-decoder neural network for intermodal image appearance prediction to match image features intermodally in an indirect manner.

\begin{figure*}[!ht]
\normalsize
      \centering
      \includegraphics[scale=0.064]{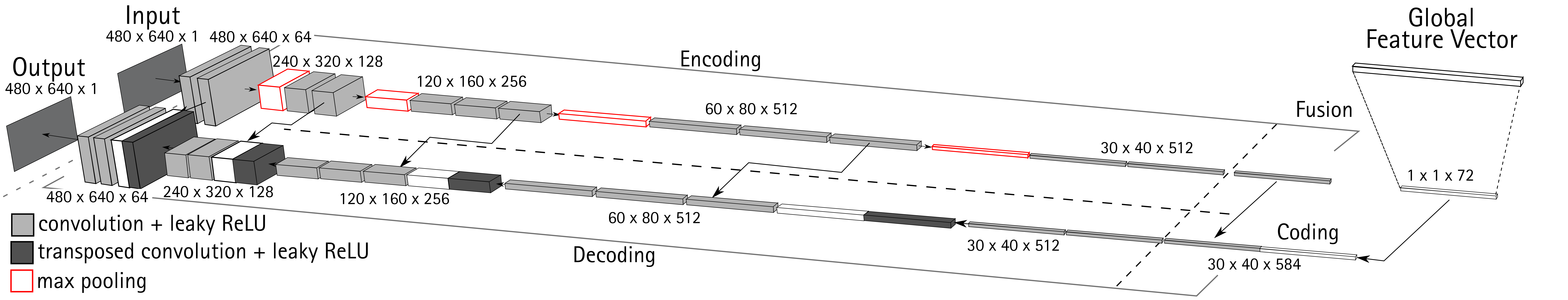}
      \caption{Network architecture of the convolutional encoder-decoder used for the intermodal image prediction step of Se-DIFT.}
      \label{networkarchitecture}
\hrulefill
\vspace*{4pt}
\end{figure*}

In contrast to the related publications presented in the previous section, the prediction of image appearances in varying imaging modalities has not only to be plausible in terms of human perception but also to be sufficiently correct objectively. While there is an almost unlimited number of training data for image colorization, which can be generated by converting any color image to grayscale, spatial aligned multimodal images are only available in a limited number. Additionally, intermodal appearance prediction is like color prediction inherently ambiguous. Whereas for colorization multiple colors can be plausible (a flower can be yellow or red), for thermal imaging, different temperatures (and therefore thermal radiation) can be plausible too: The engine of a parked car can either be hot or cold depending on its recent usage, which may not be inferred based on the input image alone.\\
Since intermodal image prediction does not only depend on information which can be inferred based on the input image such as an object's class, material or surface properties, additional information need to be provided externally. Accordingly, we use a modality-specific global feature vector which is concatenated to the coding layer of the autoencoder. 
In the case of thermal imaging, the temperature history and the incidence of the sun's radiation of a scene cannot be extracted from the input image but may be useful information for the accurate prediction of the appearance of a scene in the thermal spectrum.\\ 
The intermodal prediction can be formalized similar to \cite{Zhang2016} as follows: Given an $n$-channel input image of one modality $\textbf{X}\in \mathbb{R}^{H \times W \times n}$ with height $H$ and width $W$, we want to learn a mapping $\widehat{\textbf{Y}}=G(\textbf{X})$ to predict an $m$-channel output image of a varying modality $\textbf{Y}\in \mathbb{R}^{H \times W \times m}$. The symbol $\widehat{\textbf{.}}$ is used to distinguish between the intermodal prediction and the ground truth image used for training.

Subsequently, because the predicted image has a different image noise characteristic than the reference image, which might make feature matching harder, a bilateral filter \cite{Tomasi1998} is applied to adjust the RGB and the thermal images. Afterward, we detect image features to identify corresponding image regions using traditional feature descriptors. All stages of our approach are described in detail subsequently.

\subsection{Intermodal Image Prediction}\label{networkarchitecturesection}
\noindent For intermodal image appearance prediction, we use the symmetric convolutional encoder-decoder neural network as shown in Fig. \ref{networkarchitecture}. An overview of the network configuration is given in Table \ref{layeroverview}. The network is designed to be trained in an end-to-end fashion using spatially aligned multimodal images. The input of the network are fixed-size, mean normalized $480 \times 640$ grayscale ($n=1$) images. We chose a stacked autoencoder architecture to learn a representation (coding) of the scene depicted in the input image, to subsequently generate an image of the same scene in a different imaging modality. All convolutional layers of the network use same padding and a stride of one with a height and width of the 2D convolution window of three. The spatial dimension of the layers decreases during encoding due to max pooling operations, whereas the depth increases due to convolution layers. On the one hand, with each decrease of spatial dimension, the receptive field of the next convolution increases which is required for the extraction of high-level information. On the other hand, the decrease of spatial dimension causes a loss of low-level information which are needed to generate sharp output images. To make low-level features available for the decoder, we apply skip connection layers between the corresponding stages of the encoder and the decoder. The skip connection layers concatenate the activations of the encoders convolution layers with the activations of the decoder (as shown in Fig. \ref{networkarchitecture}). 
Necessary information for intermodal image prediction, which is not expected to be inferred based on the input image alone, is added to the coding layer by concatenating them as a global feature vector before decoding (similar to \cite{Iizuka2016}): 
\begin{align}
\textbf{y}^{\text{[fusion]}}_{u,v} = \sigma(\textbf{W}^{\text{[fusion]}}\begin{bmatrix}\textbf{a}^{\text{[coding]}}_{u,v}\\\textbf{y}^{\text{[global]}}\end{bmatrix}+\textbf{b}^{\text{[fusion]}}),
\end{align}
where $u$ and $v$ are the coordinates of the coding layer activation $\textbf{a}^{\text{[coding]}}$ and the global feature vector $\textbf{y}^{\text{[global]}}$, the activation function $\sigma$, the weight matrix of the fusion layer $\textbf{W}^{\text{[fusion]}}$ and the bias vector $\textbf{b}^{\text{[fusion]}}$.
While dimensionality reduction in the encoder is performed using max pooling layer, the decoding stage is based on transposed convolutions for upsampling.\\
To make the network trainable with the typically small quantity of available spatially corresponding multimodal training data, we use the first 13 layer of the VGG16 network \cite{vgg16} to design our encoder stage to be able to apply transfer learning with pretrained weights. In contrast to the original VGG16 architecture, we use Leaky Rectified Linear Units (leaky ReLUs) as activation function $\sigma$ to avoid dying ReLUs. We preferred using leaky ReLU compared to Exponential Linear Units (ELUs) for reasons of runtime performance. For the final output layer, we employ the hyperbolic tangent function. To avoid overfitting, we apply L2-Regularization \cite{Ng2004} and Dropout \cite{Hinton2006}. The dropout rate of the skip connection layers is chosen to be much higher than the rate for the rest of the network. Even Batch Normalization (BN) has shown to speed up learning and allows the training of deep neural networks, we forego to normalize the activations in favor to be able to use the pretrained VGG16 weights (which have been trained without BN).
\begin{table*}[t]
\caption{Overview of the networks encoder and decoder layer.}\label{layeroverview}
\centering
\begin{tabular}{p{2cm}|p{0.15cm}|p{1.25cm}|p{0.15cm}|p{1.25cm}|p{0.15cm}|p{1.25cm}|p{0.15cm}|p{1.25cm}|p{0.15cm}|p{1.25cm}|p{0.15cm}|p{2cm}}
\hlineplus
\multicolumn{13}{c}{\textit{Encoder}}\\
\hlineplus
\hline
 \textbf{Input} ($480 \times 640 \times 1$) & \rotatebox[origin=c]{90}{\begin{tabular}{c}\end{tabular} }& \begin{tabular}{c} conv3-64\\ conv3-64 \end{tabular} & \rotatebox[origin=c]{90}{\begin{tabular}{c}maxpool\end{tabular} } & \begin{tabular}{c} conv3-128\\ conv3-128 \end{tabular} & \rotatebox[origin=c]{90}{maxpool} & \begin{tabular}{c}conv3-256\\ conv3-256\\ conv3-256\end{tabular} & \rotatebox[origin=c]{90}{maxpool } & \begin{tabular}{c}conv3-512\\ conv3-512\\ conv3-512\end{tabular} & \rotatebox[origin=c]{90}{maxpool } & \begin{tabular}{c}conv3-512\\ conv3-512\\ conv3-512\end{tabular} & \rotatebox[origin=c]{90}{maxpool } & \textbf{Coding} ($15 \times 20 \times 512$)\\ \\
\hlineplus
\multicolumn{13}{c}{\textit{Decoder}}\\
\hlineplus
\hline
\textbf{Coding + global feature vector} ($15 \times 20 \times 584$) & \rotatebox[origin=c]{90}{\begin{tabular}{c}t.-conv\end{tabular} }& \begin{tabular}{c}\\ conv3-512\\ conv3-512\\ conv3-512\end{tabular} & \rotatebox[origin=c]{90}{t.-conv} & \begin{tabular}{c}\\conv3-512\\ conv3-512\\ conv3-512\end{tabular} & \rotatebox[origin=c]{90}{t.-conv} & \begin{tabular}{c}\\conv3-256\\ conv3-256\\ conv3-256\end{tabular} & \rotatebox[origin=c]{90}{t.-conv} & \begin{tabular}{c}\\conv3-128\\ conv3-128\end{tabular} & \rotatebox[origin=c]{90}{t.-conv} & \begin{tabular}{c}\\conv3-64\\ conv3-64 \end{tabular} & \rotatebox[origin=c]{90}{} & \textbf{Output} ($480 \times 640 \times 1$)\\ \hlineplus
\end{tabular}
\end{table*}
As objective function, we use the Euclidean loss between spatially aligned ground truth images $\textbf{Y}$ and the predicted image $\widehat{\textbf{Y}}=G(\textbf{X})$:
\begin{align}
\mathcal{L}_{L1}(G)= \mathds{E}_{\textbf{X},\textbf{Y}}[||\textbf{Y}-G(\textbf{X})||_1]
\end{align}
Furthermore, we also train the network with a cGAN as loss function which is defined as follows (according to the combined L1-cGAN loss used in \cite{px2px}):
\begin{align}
\mathcal{L}_{cGAN}(G,D) = &\mathds{E}_{\textbf{X},\textbf{Y}}[log~D(\textbf{X},\textbf{Y})]+\\
&\mathds{E}_{\textbf{X}}[log~(1-D(\textbf{X},G(\textbf{X}))]
\end{align}
The design of the discriminator $D$ is chosen to be identical to the experiments in \cite{px2px} to make the results comparable.
The final objective results in a weighted combination of the L1 and cGAN loss:
\begin{align}
G* = arg~\underset{G}{min}~\underset{D}{max}~\alpha~\mathcal{L}_{L1}(G) + \beta~\mathcal{L}_{cGAN}(G,D).
\end{align}
With $\beta=0$ only the L1 loss is optimized. To optimize our network, we use the Adam optimization algorithm \cite{adam}.

\subsection{Global Feature Vector}
\noindent To design a global feature vector which adds information to the decoding stage, which can not be extracted based on the image alone, an understanding of the specific imaging process is essential. The imaging process of a thermal camera is described in detail in \cite{thermalimaging}. At his point, we will focus on the different factors which determine the pixel response of a thermal camera. Thermal imaging visualizes thermal radiation in the range of approx. 0.9 to 14 $\mu m$. All objects with a temperature above 0 K emit thermal radiation. Planck's law describes the amount of radiation that is emitted at a specific wavelength and temperature $T$ by a black body:
\begin{align}
I_{obj}(\lambda, T) = \frac{2 \pi h c^2}{\lambda^5 (e^{hc/\lambda k_BT}-1)},\label{PlanckEq}
\end{align}
where $\lambda$ is the wavelength, $h$ the Planck constant, $c$ the speed of light and $k_B$ the Boltzmann constant \cite{Gade2014}. Most objects differ from the assumption to be a black body and are called grey bodies. To make (\ref{PlanckEq}) valid for grey bodies also, a scale factor which depends on the object's material but also on the nature of its surface, the emissivity $\varepsilon$, has been introduced. $\varepsilon$~is in the range between 0 and 1. Besides emission from the object, there is reflected emission from ambient sources as well as from the atmosphere. The pixel response of a thermal camera $S$ is based on the sum of the thermal radiation over the range of wavelengths the camera is sensitive for:
\begin{align}\label{thermalpixelresponse}
S=\int_{\lambda}\ \varepsilon\ I_{obj}(\lambda, T_{obj}) + (1-\varepsilon)I_{amb}(\lambda, T_{amb})\ d\lambda,
\end{align}
whereas (\ref{thermalpixelresponse}) neglects the influence of atmospheric gases between the camera and the object's surface properties. $\varepsilon$ is the emissivity of the object and $I_{amb}$ the reflected ambient radiation.
Besides of an object's temperature $T$, the response of the thermal camera is mainly determined by its emissivity $\varepsilon$ which not only depends on the material but also on the nature of its surface. Although the emissivity is almost constant for most viewing angles, it decreases or increases (dependent whether the material is a dielectric or not) at relatively high viewing angles. To predict an object's appearance in a thermal image, the object's temperature, material, surface properties, and viewing direction needs to be estimated from the RGB image or has to be provided externally.\\ 
Whereas material, surface properties, and viewing direction are expected to be estimable based on the RGB image, additional information to estimate the temperature needs to be provided externally. The temperature of an object is mainly influenced by the object's mass, its specific heat capacity and the amount of energy supplied to the object. The mass of an object and the thermal capacity are material specific and therefore might be estimated based on the RGB image also. For the supplied energy, we neglect active heating and only consider the energy supplied by the environment. For this reason, we provide the temperature history for the previous 72 hours as global feature vector using an external weather database. The number of hours of sunshine, as well as the sun's intensity, might also be useful. For reasons of simplicity, we assume that a longer period of sunshine results in a higher overall temperature and therefore does not need to be provided as well.
\subsection{Training Data Generation}
\noindent Large quantities of spatially aligning multimodal images are required to train the presented network. Multimodal cameras typically distinguish in intrinsic parameters as well as camera poses and therefore the resulting images need to be registered first. To reduce parallax, we choose a small baseline multimodal setup as described in \cite{Kleinschmidt2017}. Based on the assumption that for the selected camera setup the depth of the scene is significantly larger than the distance between the cameras, the resulting disparity can be assumed to be almost constant for all pixels and therefore can be corrected mathematically.\\ 
To decrease the number of necessary multimodal training images, we apply transfer learning using a pretrained VGG16 network to initialize the encoder part of the network. Besides, we apply data augmentation techniques such as random rotations, shifting, scaling, and flipping to the corresponding image pairs of our training set.
\section{EXPERIMENTS}
\subsection{Setup}
\noindent To train our network, we generated 8516 individual, spatially aligning image pairs using a \textit{FLIR One Pro} camera providing RGB and thermal input images in a resolution of $1440 \times 1.080$ and $160 \times 120$ pixel. The camera has been connected to a Motorola X4 smartphone, on which the images are spatially aligned subsequently. The resulting dataset includes pictures of urban and natural environments as well as different objects as images of parked cars. The pictures have been taken across all seasons including winter and summer but only at day time. The spatially aligned RGB and thermal images have a size of $480 \times 640$ pixel. The upscaling of the thermal images results in more blur compared to the RGB images.
To evaluate Se-DIFT, the resulting dataset has been randomly split into $90\%$ training, $5\%$ validation and $5\%$ test set. All experiments in this section are performed on the resulting 427 images of the test set which have been excluded from training as well as from network desgin optimization process. The test set contains 222 images of urban as well as 88 images of natural environments and 116 images of objects. Considering, that every image provides at least 50 sparse feature points, the number of features is significantly higher than the number of input images.\\ 
The network is trained with a learning rate of $0.0001$ on a \textit{NVIDIA Titan V}. The training ends after 200 epochs or when the validation error did not increase for $10$ epochs.\\
To investigate the transferability of different feature detectors and descriptors using Se-DIFT, we evaluate SIFT, SURF, and ORB features. To provide ground truth positions of corresponding image features, features detected in the predicted thermal images are matched against features detected in the real thermal image. Consequently, an ideal feature pair would have an Euclidean distance of $0\ px$. 
Considering that both input images are spatially aligned according to Eq. \ref{mapping} using a homography, the remaining error after registration can be computed according Eq. \ref{remainingerror}:
\begin{align}\label{mapping}
_{(C_2)}\textbf{x}_p \mapsto _{(C_1)}\textbf{x}_p,~ {}_{(C_1)}\widetilde{\textbf{x}}_p' = {}^{(C1)}\textbf{H}_{(C2)} {}_{(C_2)}\overline{\textbf{x}}_p.
\end{align}
Whereas ${}^{(C1)}\textbf{H}_{(C2)}$ can be computed according to Eq.~\ref{projcorrectiona} using the intrinsic and extrinsic camera calibration.
\begin{align}\label{projcorrectiona}
{}^{(C1)}\textbf{H}_{(C2)} &= {}_{(C1)}\textbf{M}~ {}^{(C1)}\textbf{R}~_{(C2)}~ {}_{(C2)}\textbf{M}^{-1}.
\end{align}
Considering, that there is no rotation between the cameras, the remaining error in pixel can be computed according to Eq.~\ref{remainingerror}:
\begin{align}\label{remainingerror}
{}_{(C1)}e &= \vert \vert {}_{(C_1)}\textbf{x}_p - {}_{(C_1)}\textbf{x}_p' \vert \vert\\
&= \biggl| \biggl| \left(\begin{array}{c} \frac{\mathrm{{}_{(C1)}f_x}\,\left(\mathrm{{}_{(C1)}X_p}\,\mathrm{{}^{(C1)}t_z{}_{(C2)}}-\mathrm{{}_{(C1)}Z_p}\,\mathrm{{}^{(C1)}t_x{}_{(C2)}}\right)}{\mathrm{{}_{(C1)}Z_p}\,\left(\mathrm{{}_{(C1)}Z_p}+\mathrm{{}^{(C1)}t_z{}_{(C2)}}\right)}\\ \frac{\mathrm{{}_{(C1)}f_y}_{1}\,\left(\mathrm{{}_{(C1)}Y_p}\,\mathrm{{}^{(C1)}t_z{}_{(C2)}}-\mathrm{{}_{(C1)}Z_p}\,\mathrm{{}^{(C1)}t_y{}_{(C2)}}\right)}{\mathrm{{}_{(C1)}Z_p}\,\left(\mathrm{{}_{(C1)}Z_p}+\mathrm{{}^{(C1)}t_z{}_{(C2)}}\right)} \end{array}\right) \biggl| \biggl|.
\end{align}
With a baselength between the cameras of $1~cm$ and a minimal distance to the environment of more than $2.5~m$, the remaining registration error is assumed to be less than $5~px$ according to Eq.~\ref{remainingerror} (assuming an accurate intrinsic and extrinsic calibration). For this reason, image feature matches are accepted as correct if the Euclidean distance between the matched and the associated ground truth feature points of the spatially aligned image is below the expected error.

\subsection{Appearance Prediction}
To evaluate the intermodal prediction of our network, we make predictions for the images of the test set and compare them against the registered ground truth images by computing L1 norm. The validate the improvement of using the coding augmentation presented in this paper, we compare the network against its non-augmented version. Furthermore, we compare our results against a version trained with cGAN and a weighted combination of L1 and cGAN loss (as used in \cite{px2px}).

\subsection{Feature Matching}
\begin{figure*}[t!]
\normalsize
\centering
  \subfloat[input image]{%
       \includegraphics[width=0.16\linewidth]{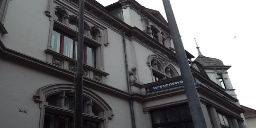}\label{results_1a}}
  \subfloat[intermodal prediction]{%
        \includegraphics[width=0.16\linewidth]{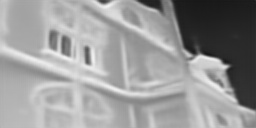}\label{results_1b}}
      \subfloat[ground truth]{%
        \includegraphics[width=0.16\linewidth]{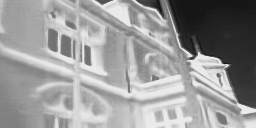}\label{results_1c}}
        \quad
      \subfloat[input image]{%
       \includegraphics[width=0.16\linewidth]{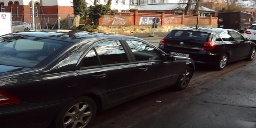}\label{results_2a}}
  \subfloat[intermodal prediction]{%
        \includegraphics[width=0.16\linewidth]{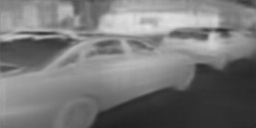}\label{results_2b}}
      \subfloat[ground truth]{%
        \includegraphics[width=0.16\linewidth]{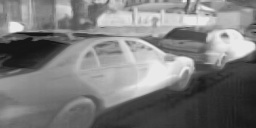}\label{results_2c}}\\
      \subfloat[input image]{%
       \includegraphics[width=0.16\linewidth]{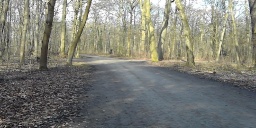}\label{results_3a}}
  \subfloat[intermodal prediction]{%
        \includegraphics[width=0.16\linewidth]{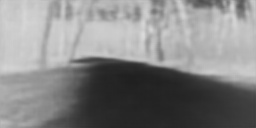}\label{results_3b}}
      \subfloat[ground truth]{%
        \includegraphics[width=0.16\linewidth]{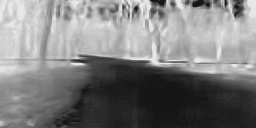}\label{results_3c}} \quad
      \subfloat[input image]{%
       \includegraphics[width=0.16\linewidth]{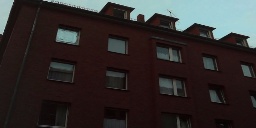}\label{results_4a}}
  \subfloat[intermodal prediction]{%
        \includegraphics[width=0.16\linewidth]{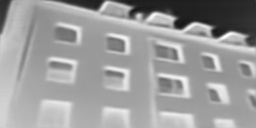}\label{results_4b}}
      \subfloat[ground truth]{%
        \includegraphics[width=0.16\linewidth]{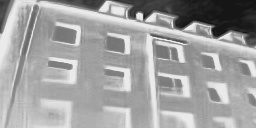}\label{results_4c}}\\
      \subfloat[input image]{%
       \includegraphics[width=0.16\linewidth]{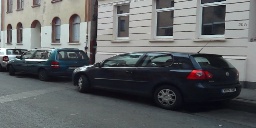}}
    \label{results_5a}
  \subfloat[intermodal prediction]{%
        \includegraphics[width=0.16\linewidth]{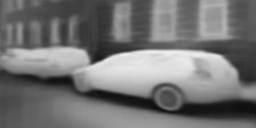}\label{results_5b}}
      \subfloat[ground truth]{%
        \includegraphics[width=0.16\linewidth]{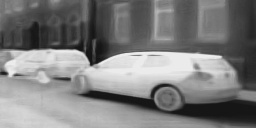}\label{results_5c}} \quad 
      \subfloat[input image]{%
       \includegraphics[width=0.16\linewidth]{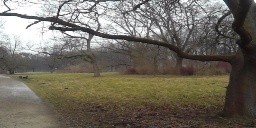}\label{results_6a}}
  \subfloat[intermodal prediction]{%
        \includegraphics[width=0.16\linewidth]{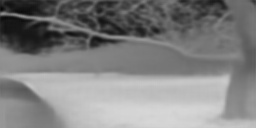}\label{results_6b}}
      \subfloat[ground truth]{%
        \includegraphics[width=0.16\linewidth]{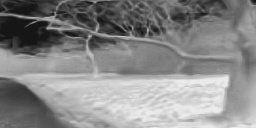}\label{results_6c}}\\
  \caption{Intermodal predictions of the test set generated using Se-DIFT.}
  \label{figUnknownBackgroundInformation} 
\vspace*{4pt}
\end{figure*}
\noindent To proof the capabilities of Se-DIFT, we evaluate SIFT, SURF, and ORB feature descriptors. The descriptors are matched using brute force matching with L2 norm as distance measurement for SIFT and SURF, and the Hamming distance for ORB descriptors. As proposed by D. Lowe in \cite{SIFT}, we perform the ratio test on the two best matches for each feature. Features are sorted by their significance and a maximum number of 100 features is kept for feature matching. The matching threshold are varyied to generate receiver operating characteristics (ROC curves) for all feature network-combinations. The ROC curve plots the true positive rate against the false positive rate. To evaluate the performance, we then compute the area under curve (AUC) and the equal error rates (EER) for all ROC curves. An AUC close to one and an EER close to zero indicate an ideal performance of the respective feature-network combination.
\section{RESULTS}
\newcommand{\VGGUNetRGBTHERMALLcloss}{0.16}
\newcommand{\VGGUNetRGBTHERMALLbloss}{0.14}
\newcommand{\VGGUNetRGBTHERMALLnloss}{0.13}
\newcommand{\VGGUNetRGBTHERMALLtotalloss}{0.14}
\newcommand{\VGGUNetRGBTHERMALLparametercount}{41857666}

\newcommand{\VGGUNetTHERMALRGBLcloss}{0.17}
\newcommand{\VGGUNetTHERMALRGBLbloss}{0.15}
\newcommand{\VGGUNetTHERMALRGBLnloss}{0.15}
\newcommand{\VGGUNetTHERMALRGBLtotalloss}{0.16}
\newcommand{\VGGUNetTHERMALRGBLparametercount}{41857666}

\newcommand{\VGGUNetWeatherRGBTHERMALLcloss}{0.14}
\newcommand{\VGGUNetWeatherRGBTHERMALLbloss}{0.13}
\newcommand{\VGGUNetWeatherRGBTHERMALLnloss}{0.12}
\newcommand{\VGGUNetWeatherRGBTHERMALLtotalloss}{0.13}
\newcommand{\VGGUNetWeatherRGBTHERMALLparametercount}{42336898}

\newcommand{\VGGUNetWeatherTHERMALRGBLcloss}{0.16}
\newcommand{\VGGUNetWeatherTHERMALRGBLbloss}{0.15}
\newcommand{\VGGUNetWeatherTHERMALRGBLnloss}{0.14}
\newcommand{\VGGUNetWeatherTHERMALRGBLtotalloss}{0.15}
\newcommand{\VGGUNetWeatherTHERMALRGBLparametercount}{42336898}

\newcommand{\VGGUNetRGBTHERMALcGANcloss}{0.19}
\newcommand{\VGGUNetRGBTHERMALcGANbloss}{0.16}
\newcommand{\VGGUNetRGBTHERMALcGANnloss}{0.15}
\newcommand{\VGGUNetRGBTHERMALcGANtotalloss}{0.17}
\newcommand{\VGGUNetRGBTHERMALcGANparametercount}{45486596}

\newcommand{\VGGUNetTHERMALRGBcGANcloss}{0.18}
\newcommand{\VGGUNetTHERMALRGBcGANbloss}{0.16}
\newcommand{\VGGUNetTHERMALRGBcGANnloss}{0.16}
\newcommand{\VGGUNetTHERMALRGBcGANtotalloss}{0.17}
\newcommand{\VGGUNetTHERMALRGBcGANparametercount}{45486596}


\newcommand{\VGGUNetWeatherRGBTHERMALcGANcloss}{0.21}
\newcommand{\VGGUNetWeatherRGBTHERMALcGANbloss}{0.18}
\newcommand{\VGGUNetWeatherRGBTHERMALcGANnloss}{0.19}
\newcommand{\VGGUNetWeatherRGBTHERMALcGANtotalloss}{0.19}
\newcommand{\VGGUNetWeatherRGBTHERMALcGANparametercount}{45965828}

\newcommand{\VGGUNetWeatherTHERMALRGBcGANcloss}{0.23}
\newcommand{\VGGUNetWeatherTHERMALRGBcGANbloss}{0.24}
\newcommand{\VGGUNetWeatherTHERMALRGBcGANnloss}{0.18}
\newcommand{\VGGUNetWeatherTHERMALRGBcGANtotalloss}{0.22}
\newcommand{\VGGUNetWeatherTHERMALRGBcGANparametercount}{45965828}

\newcommand{\VGGUNetRGBTHERMALLcGANcloss}{0.19}
\newcommand{\VGGUNetRGBTHERMALLcGANbloss}{0.16}
\newcommand{\VGGUNetRGBTHERMALLcGANnloss}{0.15}
\newcommand{\VGGUNetRGBTHERMALLcGANtotalloss}{0.17}
\newcommand{\VGGUNetRGBTHERMALLcGANparametercount}{45486596}

\newcommand{\VGGUNetTHERMALRGBLcGANcloss}{0.18}
\newcommand{\VGGUNetTHERMALRGBLcGANbloss}{0.16}
\newcommand{\VGGUNetTHERMALRGBLcGANnloss}{0.16}
\newcommand{\VGGUNetTHERMALRGBLcGANtotalloss}{0.17}
\newcommand{\VGGUNetTHERMALRGBLcGANparametercount}{45486596}

\newcommand{\VGGUNetWeatherRGBTHERMALLcGANcloss}{0.15}
\newcommand{\VGGUNetWeatherRGBTHERMALLcGANbloss}{0.13}
\newcommand{\VGGUNetWeatherRGBTHERMALLcGANnloss}{0.12}
\newcommand{\VGGUNetWeatherRGBTHERMALLcGANtotalloss}{0.14}
\newcommand{\VGGUNetWeatherRGBTHERMALLcGANparametercount}{45965828}

\newcommand{\VGGUNetWeatherTHERMALRGBLcGANcloss}{0.17}
\newcommand{\VGGUNetWeatherTHERMALRGBLcGANbloss}{0.15}
\newcommand{\VGGUNetWeatherTHERMALRGBLcGANnloss}{0.15}
\newcommand{\VGGUNetWeatherTHERMALRGBLcGANtotalloss}{0.15}
\newcommand{\VGGUNetWeatherTHERMALRGBLcGANparametercount}{45965828}

\newcommand{\simpleUNetRGBTHERMALLcloss}{Atbc}
\newcommand{\simpleUNetRGBTHERMALLbloss}{Btbc}
\newcommand{\simpleUNetRGBTHERMALLnloss}{Ctbc}
\newcommand{\simpleUNetRGBTHERMALLtotalloss}{Dtbc}
\newcommand{\simpleUNetRGBTHERMALLparametercount}{-}

\newcommand{\simpleUNetTHERMALRGBLcloss}{Atbc}
\newcommand{\simpleUNetTHERMALRGBLbloss}{Btbc}
\newcommand{\simpleUNetTHERMALRGBLnloss}{Ctbc}
\newcommand{\simpleUNetTHERMALRGBLtotalloss}{Dtbc}
\newcommand{\simpleUNetTHERMALRGBLparametercount}{-}

\newcommand{\simpleUNetRGBTHERMALcGANcloss}{Etbc}
\newcommand{\simpleUNetRGBTHERMALcGANbloss}{Ftbc}
\newcommand{\simpleUNetRGBTHERMALcGANnloss}{Gtbc}
\newcommand{\simpleUNetRGBTHERMALcGANtotalloss}{Htbc}
\newcommand{\simpleUNetRGBTHERMALcGANparametercount}{-}

\newcommand{\simpleUNetTHERMALRGBcGANcloss}{Etbc}
\newcommand{\simpleUNetTHERMALRGBcGANbloss}{Ftbc}
\newcommand{\simpleUNetTHERMALRGBcGANnloss}{Gtbc}
\newcommand{\simpleUNetTHERMALRGBcGANtotalloss}{Htbc}
\newcommand{\simpleUNetTHERMALRGBcGANparametercount}{-}

\newcommand{\simpleUNetRGBTHERMALLcGANcloss}{Itbc}
\newcommand{\simpleUNetRGBTHERMALLcGANbloss}{Jtbc}
\newcommand{\simpleUNetRGBTHERMALLcGANnloss}{Ktbc}
\newcommand{\simpleUNetRGBTHERMALLcGANtotalloss}{Ltbc}
\newcommand{\simpleUNetRGBTHERMALLcGANparametercount}{-}

\newcommand{\simpleUNetTHERMALRGBLcGANcloss}{Itbc}
\newcommand{\simpleUNetTHERMALRGBLcGANbloss}{Jtbc}
\newcommand{\simpleUNetTHERMALRGBLcGANnloss}{Ktbc}
\newcommand{\simpleUNetTHERMALRGBLcGANtotalloss}{Ltbc}
\newcommand{\simpleUNetTHERMALRGBLcGANparametercount}{-}

\newcommand{\simpleUNetRGBTHERMALcycleGANcloss}{Mtbc}
\newcommand{\simpleUNetRGBTHERMALcycleGANbloss}{Ntbc}
\newcommand{\simpleUNetRGBTHERMALcycleGANnloss}{Otbc}
\newcommand{\simpleUNetRGBTHERMALcycleGANtotalloss}{Ptbc}
\newcommand{\simpleUNetRGBTHERMALcycleGANparametercount}{-}

\newcommand{\simpleUNetTHERMALRGBcycleGANcloss}{Mtbc}
\newcommand{\simpleUNetTHERMALRGBcycleGANbloss}{Ntbc}
\newcommand{\simpleUNetTHERMALRGBcycleGANnloss}{Otbc}
\newcommand{\simpleUNetTHERMALRGBcycleGANtotalloss}{Ptbc}
\newcommand{\simpleUNetTHERMALRGBcycleGANparametercount}{-}

\begin{table*}[t]
\small
\centering
\caption{Reconstruction losses for the different network architectures and datasets predicting thermal images based on RGB images as input.}\label{reconstructionresults}
\begin{tabular}{llllll|llll}
\hlineplus
\multicolumn{2}{c}{\multirow{3}{*}{\textbf{Network Architecture and Loss}}} & \multicolumn{8}{c}{\textbf{Per-Pixel-loss}}                         \\
\cline{3-10}
&  & \multicolumn{4}{c}{\textbf{RGB-Thermal}} & \multicolumn{4}{c}{\textbf{Thermal-RGB}}                        \\
\cline{3-10}
           & & Objects & Buildings & Nature & All & Objects & Buildings & Nature & All\\
                    \hlineplus
                    & Regular, \textit{L1} & \VGGUNetRGBTHERMALLcloss & \VGGUNetRGBTHERMALLnloss & \VGGUNetRGBTHERMALLbloss & \VGGUNetRGBTHERMALLtotalloss & \VGGUNetTHERMALRGBLcloss & \VGGUNetTHERMALRGBLnloss & \VGGUNetTHERMALRGBLbloss & \VGGUNetTHERMALRGBLtotalloss \\
                    & \textbf{Augmented, \textit{L1} \textit{(proposed)}} &  \textbf{\VGGUNetWeatherRGBTHERMALLcloss} & \textbf{\VGGUNetWeatherRGBTHERMALLbloss} & \textbf{\VGGUNetWeatherRGBTHERMALLnloss} &  \textbf{\VGGUNetWeatherRGBTHERMALLtotalloss} & \textbf{\VGGUNetWeatherTHERMALRGBLcloss} & \textbf{\VGGUNetWeatherTHERMALRGBLbloss} & \textbf{\VGGUNetWeatherTHERMALRGBLnloss} &  \textbf{\VGGUNetWeatherTHERMALRGBLtotalloss} \\
                    & Regular, \textit{L1+cGAN} & \VGGUNetRGBTHERMALLcGANcloss & \VGGUNetRGBTHERMALLcGANbloss & \VGGUNetRGBTHERMALLcGANnloss & \VGGUNetRGBTHERMALLcGANtotalloss  &  \VGGUNetTHERMALRGBLcGANcloss & \VGGUNetTHERMALRGBLcGANbloss & \VGGUNetTHERMALRGBLcGANnloss & \VGGUNetTHERMALRGBLcGANtotalloss\\
                    & Augmented, \textit{L1+cGAN} & \VGGUNetWeatherRGBTHERMALLcGANcloss & \VGGUNetWeatherRGBTHERMALLcGANbloss & \VGGUNetWeatherRGBTHERMALLcGANnloss & \VGGUNetWeatherRGBTHERMALLcGANtotalloss  & \VGGUNetWeatherTHERMALRGBLcGANcloss & \VGGUNetWeatherTHERMALRGBLcGANbloss & \VGGUNetWeatherTHERMALRGBLcGANnloss & \VGGUNetWeatherTHERMALRGBLcGANtotalloss\\
                  \hlineplus
\end{tabular}
\end{table*}

\subsection{Appearance Prediction}
\noindent In the following, the L1 versions of the networks are trained with $\alpha=100$ and $\beta=0$, the L1-cGAN versions are trained with $\alpha=100$ and $\beta=1$. The parameters are chosen this way to keep the result comparable to \cite{px2px}. Fig. \ref{results_1a} to \ref{results_6c} show results of the intermodal prediction step of Se-DIFT using the network architecture shown in Fig.~\ref{networkarchitecture} with the proposed global feature vector, trained with $\alpha=100$ and $\alpha=0$ (L1 loss only). As shown in Table \ref{reconstructionresults}, our approach achieves a lower L1 error compared to the non augmented network architecture. Compared to the ground truth images, the direction of gradients have been predicted correctly for most image segments. Although skip connections have been added to the network, the predicted images provide fewer details as a consequence of the max pooling operations of the network's encoder. Therefore, image features on fine structures are not expected to be transferable using the current network architecture. In general, predicting the thermal image based on the RGB image results in a lower L1 error than the prediction of the RGB image based on the thermal image.\\
The network trained with a L1 loss in combination with a cGAN was able to genereate sharper contours. Eventhough, the generated structures deviate more strongly from the real structures and therefore result in a higher L1 error leading to a worse feature matching performance. In general, the intermodal prediction fails in case of unknown background information: As shown in Fig. \ref{figUnknownBackgroundInformation}, the algorithm failed in predicting the appearance of the car correctly, not knowing that it has been recently moved and therefore has a higher level of thermal radiation in the area of the engine bonnet than predicted. This can be reasoned by the fact, that the state of the engine could neither be inferred by the input image nor by the provided global feature vector. Similar predictions can be partially observed in the urban dataset, because there is no information on whether an apartment is heated or not. Therefore, the amount of radiation emitted through the windows is unknown. 
\begin{figure}[t!]
\centering
  \subfloat[Intermodal prediction]{%
       \includegraphics[width=0.49\linewidth]{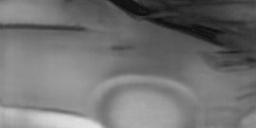}}
    \label{1a}\hfill
  \subfloat[Ground truth image]{%
        \includegraphics[width=0.49\linewidth]{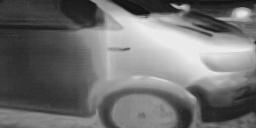}}
    \label{1b}\\
  \caption{Limitations of the intermodal prediciton used for Se-DIFT. The history of the object could not be inferred by the input RGB image nor the global feature vector and therefore is not predicted correctly.}
  \label{figUnknownBackgroundInformation} 
\end{figure}

\subsection{Feature Matching}
\noindent For evaluation, SIFT, SURF, and ORB features detected in the predicted images are matched against features detected in the ground truth images. Table \ref{sediftresultsA} and \ref{sediftresultsB} show the resulting $AUC$ and $EERs$ for the different image feature and network architecture combinations.

\begin{table*}[t!]
\begin{center}
\begin{tabular}{cc}

\begin{minipage}{.5\linewidth}
\caption{AUC for different feature descriptors and prediction networks}
\centering
\begin{tabular}{ccp{1cm}p{1cm}}
\hlineplus
& & \multicolumn{2}{c}{\textbf{Area Under Curve (AUC)}}\\
\hlineplus
& \textbf{Network} & RGB-Thermal & Thermal-RGB \\
\hlineplus
\multirow{5}{*}{\rotatebox[origin=c]{90}{\textbf{SIFT}}} & \textit{no prediction} & 0.15 & 0.16\\
 & \textit{Regular, L1} & 0.61 & 0.58\\
 & \textbf{Augmented, L1} & \textbf{0.67} & \textbf{0.60}\\
 & \textit{Regular, L1+cGAN} & 0.49 & 0.65\\
 & \textit{Augmented, L1+cGAN} & 0.42 & 0.49 \\
\hline
\multirow{5}{*}{\rotatebox[origin=c]{90}{\textbf{SURF}}} & \textit{no prediction} & 0.14 & 0.16 \\
 & \textit{Regular, L1} & 0.54 & 0.51\\
 & \textbf{Augmented, L1} & \textbf{0.60} & \textbf{0.55}\\
 & \textit{Regular, L1+cGAN} & 0.39 & 0.47\\
 & \textit{Augmented, L1+cGAN} & 0.30 & 0.33\\
\hline
\multirow{5}{*}{\rotatebox[origin=c]{90}{\textbf{ORB}}} & \textit{no prediction} & 0.18 & 0.17 \\
 & \textit{Regular, L1} & 0.45 & 0.43\\
 & \textbf{Augmented, L1} & \textbf{0.48} & \textbf{0.45}\\
 & \textit{Regular, L1+cGAN} & 0.36 & 0.33\\
 & \textit{Augmented, L1+cGAN} & 0.26 & 0.24\\
\hlineplus
\end{tabular}\label{sediftresultsA}
\end{minipage}

\begin{minipage}{.5\linewidth}
\caption{EER for different feature descriptors and prediction networks}
\centering
\begin{tabular}{ccp{1cm}p{1cm}}
\hlineplus
& & \multicolumn{2}{c}{\textbf{Equal Error Rate (EER)}}\\
\hlineplus
& \textbf{Network} & RGB-Thermal & Thermal-RGB \\
\hlineplus
\multirow{5}{*}{\rotatebox[origin=c]{90}{\textbf{SIFT}}} & \textit{no prediction} & 0.58 & 0.72\\
 & \textit{Regular, L1} & 0.40 & 0.44\\
 & \textbf{Augmented, L1} & \textbf{0.24} & \textbf{0.42}\\
 & \textit{Regular, L1+cGAN} & 0.49 & 0.37\\
 & \textit{Augmented, L1+cGAN} & 0.46 & 0.47 \\
\hline
\multirow{5}{*}{\rotatebox[origin=c]{90}{\textbf{SURF}}} & \textit{no prediction} & 0.63 & 0.71 \\
 & \textit{Regular, L1} & 0.39 & 0.44\\
 & \textbf{Augmented, L1} & \textbf{0.36} & \textbf{0.41}\\
 & \textit{Regular, L1+cGAN} & 0.46 & 0.47\\
 & \textit{Augmented, L1+cGAN} & 0.47 & 0.51\\
\hline
\multirow{5}{*}{\rotatebox[origin=c]{90}{\textbf{ORB}}} & \textit{no prediction} & 0.65 & 0.75 \\
 & \textit{Regular, L1} & 0.49 & 0.50\\
 & \textbf{Augmented, L1} & \textbf{0.49} & \textbf{0.50}\\
 & \textit{Regular, L1+cGAN} & 0.54 & 0.51\\
 & \textit{Augmented, L1+cGAN} & 0.59 & 0.56\\
\hlineplus
\end{tabular}\label{sediftresultsB}
\end{minipage}
\end{tabular}
\end{center}
\end{table*}

As the tables indicate, the feature matching using Se-DIFT in combination with the network architecture with augmented coding as proposed in this paper outperforms the other network architectures regarding the AUC and EER of the ROC curves. As the low values of the AUC and the high values for the EER of a direct matching indicate, SIFT, SURF and ORB features can not be applied for intermodal feature matching directly. Even if Se-DIFT in combination with any network architectures results in a drastically increased AUC and decreased EER compared to the direct comparison, the combination of the proposed network architecture with a L1 loss outperform the other network architectures regarding the AUC and EERs. The only exception is the AUC of the Thermal-RGB matching based on the prediction of an unaugmented network trained with a cGAN loss. In this particular measurement, the AUC is higher than the AUC of the proposed combination. Furthermore, the EER of this combination is also lower.\\
In general, the AUC for intermodal matching based on predicted thermal images is higher, and the EERs are lower, than the intermodal matching based on predicted RGB images. This is due to the fact that although RGB images provide a significant amount of relevant information for the prediction of thermal images (material, surface character, color), thermal images do not contain as much information about the representation in the visible spectrum: While material information can be easily estimated using the rich structure information of the RGB-images, the transfer from thermal to RGB images is potentially more difficult. Materials may be estimated by the thermal emissivity which, among others, depends on the temperature (and therefore on the thermal capacity of the material) but other characteristics such as surface character and color may be harder to be inferred. The worse performance predicting RGB images based on thermal images is supported by the higher L1 error for the Thermal-RGB prediction given in Table \ref{reconstructionresults}.

\section{CONCLUSION}
\noindent In this paper, we introduced Se-DIFT, a method for establishing intermodal image feature correspondences using traditional feature descriptors by predicting potential feature appearances using a convolutional encoder-decoder architecture in combination with a global feature vector which augments the encoder's coding. We demonstrated the capabilities of our approach using RGB and thermal images, which drastically distinguish in appearance and therefore are especially challenging for intermodal feature matching. Considering the generally low number of unique image features in thermal images, correct feature association is a challenging task: Many features such as transitions between different materials are ambiguous and potentially repeating over the whole image. Furthermore, most surfaces look uniformly and unstructured compared to RGB imaging. Nevertheless, we could show that by using Se-DIFT, the AUC and EER of the ROC-Curves for feature matching could be significantly improved, especially compared to a direct intermodal feature matching. Whereas the AUC increased and the EER decreased using Se-DIFT in combination with any of the presented network architectures, the best performance could be achieved using the proposed network architecture with a coding augmentation by a global feature vector as proposed and argued in this paper.\\
Besides, there are some aspects of Se-DIFT which deserve an extensive evaluation in future: Se-DIFT has only been evaluated in outdoor environments and the training dataset has been limited to buildings, objects and natural environments including trees and vegetation. To make the approach more applicable in general, the dataset variety has to be increased. Furthermore, the current evaluation of Se-DIFT only considers RGB and thermal images. Because most other imaging systems are also primarily visualizing material specific aspects of their surrounding, we expect that Se-DIFT can also be applied successfully to other imaging modalities such as hyperspectral imaging by adapting the global feature vector in the future.\\

\noindent \textbf{Acknowledgements} The Titan V used for this research was donated by the NVIDIA Corporation.

\bibliographystyle{IEEEtran}
{\small
\bibliography{biblio.bib}}

\end{document}